\documentclass[10pt,twocolumn,letterpaper]{article}

\usepackage{iccv}
\usepackage{times}
\usepackage{epsfig}
\usepackage{graphicx}
\usepackage{amsmath}
\usepackage{amssymb}

\usepackage[pagebackref=true,breaklinks=true,letterpaper=true,colorlinks,bookmarks=false]{hyperref}

\iccvfinalcopy 

\def\iccvPaperID{2610} 

\ificcvfinal\fi
\begin{document}

\title{Cascaded Context Pyramid for Full-Resolution 3D Semantic Scene Completion}
\author{
Pingping Zhang$^{\dagger}$\quad
Wei Liu$^{\ddagger}$\quad
Yinjie Lei$^{\S}$\quad
Huchuan Lu$^{\dagger}$\thanks{Prof. Lu is the corresponding author. Email: lhchuan@dlut.edu.cn.}\quad
Xiaoyun Yang$^{\star}$\\
\small$^{\dagger}$Dalian University of Technology,\quad $^{\ddagger}$University of Adelaide,\quad $^{\S}$Sichuan University,\quad $^{\star}$China Science IntelliCloud Technology Co.Ltd\\
}

\maketitle

\begin{abstract}
Semantic Scene Completion (SSC) aims to simultaneously predict the volumetric occupancy and semantic category
of a 3D scene.
It helps intelligent devices to understand and interact with the surrounding  scenes.
Due to the high-memory requirement, current methods only produce low-resolution completion predictions, and generally lose the object details.
Furthermore, they also ignore the multi-scale spatial contexts, which play a vital role for the 3D inference.
To address these issues, in this work we propose a novel deep learning framework, named Cascaded Context Pyramid Network (CCPNet), to jointly infer the occupancy and semantic labels of a volumetric 3D scene from a single depth image.
The proposed CCPNet improves the labeling coherence with a cascaded context pyramid.
Meanwhile, based on the low-level features, it progressively restores the fine-structures of objects with Guided Residual Refinement (GRR) modules.
Our proposed framework has three outstanding advantages: (1) it explicitly models the 3D spatial context for performance improvement; (2) full-resolution 3D volumes are produced with structure-preserving details; (3) light-weight models with low-memory requirements are captured with a good extensibility.
Extensive experiments demonstrate that in spite of taking a single-view depth map, our proposed framework can generate high-quality SSC results, and outperforms state-of-the-art approaches on both the synthetic SUNCG and real NYU datasets.
\end{abstract}

\section{Introduction}
Human can perceive the real-world through 3D views with partial observations.
For example, one can capture the geometry of rigid objects by only seeing the corresponding 2D images.
Thus, understanding and reconstructing a 3D scene from its partial observations is a valuable technique for many computer vision and robotic applications, such as object localization, visual reasoning and indoor navigation.
As an encouraging direction, Semantic Scene Completion (SSC) has draw more and more attentions in recent years.
It aims to simultaneously predict the volumetric occupancy and semantic category of a 3D scene.
Given a single depth image, several outstanding works~\cite{song2017semantic,guo2018view,zhang2018semantic} have been proposed for single-view SSC.
By designing 3D Convolutional Neural Networks (CNNs), these methods can automatically predict the semantic labels or complete 3D shapes of the objects in the scene.
However, it is not a trivial task to utilize 3D CNNs for the SSC task.
Vanilla 3D CNNs are locked in the cubic growth of computational and memory requirements with the increase of voxel resolution.
Thus, current methods inevitably limit the resolution of predictions and the depth of 3D CNNs, which leads to wrong labels and missing shape details in the completion results.

To achieve better SSC results, several works~\cite{guedes2018semantic,garbade2018two,liu2018see,li2019rgbd} introduce the 2D semantic segmentation as an auxiliary, which takes an additional RGB image and applies complex 2D CNNs for semantic enhancement.
These methods can fully exploit the high-resolution input, however, they ignore the 3D context information of the scene.
Thus, only based on the 2D input image, they may not infer the invisible object parts of the complex scene.
Recently, Song \emph{et al.}~\cite{song2017semantic} show that global 3D context helps the prediction of SSC.
However, the 3D CNN used in their work simply adopts the dilated convolutions~\cite{yu2016multi}, and concatenates the multi-stage features for predictions.
It only considers the global semantics, which result in low-resolution predictions, and lose the scene details.
In this work, we find that both \emph{local geometric details} and \emph{multi-scale 3D contexts} of the scene play a vital role in the SSC task.
The local geometric details help the SSC system to identify the fine-structured objects.
The multi-scale 3D contexts can enhance the spatial coherence and infer the occluded objects from the scene layout.
However, designing a framework that can efficiently integrate both characteristics is still a challenging task.

To address above problems, we propose a novel deep learning framework, named Cascaded Context Pyramid Network (CCPNet), for single depth image based SSC.
The proposed CCPNet effectively learns both local geometry details and multi-scale 3D contexts from the training dataset.
For semantic confusing objects, the CCPNet improves the prediction coherence with an effective self-cascaded context pyramid.
The self-cascaded pyramid helps the model to reduce the semantic gap of different contexts and capture the hierarchical dependencies among the objects and scenes~\cite{liu2018semantic}.
In addition, we introduce a Guided Residual Refinement (GRR) module to progressively restore the fine-structures of complex objects.
The GRR corrects the latent fitting using low-level features, and avoids the high computational cost and memory consumption of the 3D CNN.
With this module, the CCPNet can output full-resolution completion results and show much better accuracy than vanilla 3D networks.
Experimental results demonstrate that our approach outperforms other state-of-the-art methods on both synthetic and real datasets.
With only a single depth map, our method generates high-quality SSC results with much better accuracy and faster inference.

In summary, \textbf{our contributions} are three folds:
\begin{itemize}
\item
We propose a novel cascaded context pyramid network (CCPNet) for efficient 3D semantic scene completion.
The CCPNet automatically integrates both local geometric details and multi-scale 3D contexts of the scene in a self-cascaded manner.
\item
We also propose an efficient guided residual refinement (GRR) module for restoring fine-structures of objects and full-resolution predictions.
The GRR progressively refines the objects with low-level features and light-weight residual connections, improving both computational efficiency and completion accuracy.
\item
Extensive experiments on public synthetic and real benchmarks demonstrate that our proposed approach achieves superior performance over other state-of-the-art methods.
\end{itemize}
\vspace{-2mm}
\section{Related Work}
In this section, we briefly review related work on analyzing and completing a 3D scene from depth images.
%
%
For more details, we refer the readers to~\cite{ioannidou2017deep} for a survey of deep learning based 3D data processing.

\textbf{Semantic Scene Analysis.}
In recent years, many deep learning based methods have been proposed for semantic scene analysis with a depth image or RGB-D image pair.
In general, 2D image-based methods~\cite{ren2012rgb,gupta2013perceptual,wang2018depth} treat the depth image as additional information, and adopt complex 2D CNNs for semantic scene analysis tasks, \emph{e.g.}, salient object detection, semantic segmentation and scene completion.
Meanwhile, several works~\cite{gupta2014learning,gupta2015indoor,atapour2017depthcomp} extract deep features from the depth image and the RGB image separately, then fuse them for multi-mode complementarity.
Although effective, 2D image-based methods ignore the spatial occupancy of objects, and can not fully exploit the depth information.
While 3D volume-based methods usually convert the depth image into a volumetric representation, and exploit rich handcrafted 3D features~\cite{ren2016three,song2014sliding} or learned 3D CNNs~\cite{song2016deep} for detecting 3D objects.
Although existing methods can detect and segment visible 3D objects and scenes, they cannot infer the objects that are totally occluded.
Instead, our method can predict the semantic labels and 3D shapes for both visible and invisible objects.

\textbf{3D Scene Completion.}
Semantic scene completion is a fundamental task in understanding 3D scenes.
To achieve this goal, Zheng \emph{et al.}~\cite{zheng2013beyond} first complete the occluded objects with a set of pre-defined rules, and then refine the completion results by physical reasoning.
Geiger and Wang~\cite{geiger2015joint} propose a high-order graphical model to jointly reason about the layout, objects and superpixels in the scene image.
Their model leverages detailed 3D geometry of scenes, and explicitly enforces occlusion and visibility constraints.
Then, Firman \emph{et al.}~\cite{firman2016structured} utilize the random forest to infer the occluded 3D object shapes from a single depth image.
These methods are based on handcrafted features, and perform semantic scene segmentation and completion in two separate steps.
Recently, Song \emph{et al.}~\cite{song2017semantic} propose the Semantic Scene Completion Network (SSCNet) to simultaneously predict the semantic labels and volumetric occupancy of the 3D objects from a single depth image.
Although this method unifies the semantic segmentation and voxel completion, the expensive 3D CNN limits the input resolution and network depth.
Thus the SSCNet only produces low-resolution predictions and generally lacks of object details.
By combining the 2D CNN and 3D CNN, Guo and Tong~\cite{guo2018view} propose the View-Volume Network (VVNet) to efficiently reduce the computation cost and enhance the network depth.
Garbade \emph{et al.}~\cite{garbade2018two} propose a two-stream approach that jointly leverages the depth and semantic information.
They first construct an incomplete 3D semantic tensor for the inferred 2D semantic information, and then adopt a vanilla 3D CNN to infer the complete 3D semantic tensor.
Liu \emph{et al.}~\cite{liu2018see} propose a task-disentangled framework to sequentially carry out the 2D semantic segmentation, 2D-3D re-projection and 3D semantic scene completion.
However, their multi-stage method may cause the error accumulation, producing mislabeling completion results.
Similarly, Li \emph{et al.}~\cite{li2019rgbd} introduce a Dimensional Decomposition Residual Network (DDRNet) for the 3D SSC task.
Based on the factorized and dilated convolutions~\cite{chen2018deeplab}, they utilize the multi-scale feature fusion mechanism for depth and color images.
\begin{figure*}
\centering
\resizebox{0.9\textwidth}{!}
{
\begin{tabular}{@{}c@{}c@{}}
\includegraphics[width=1\linewidth,height=8.8cm]{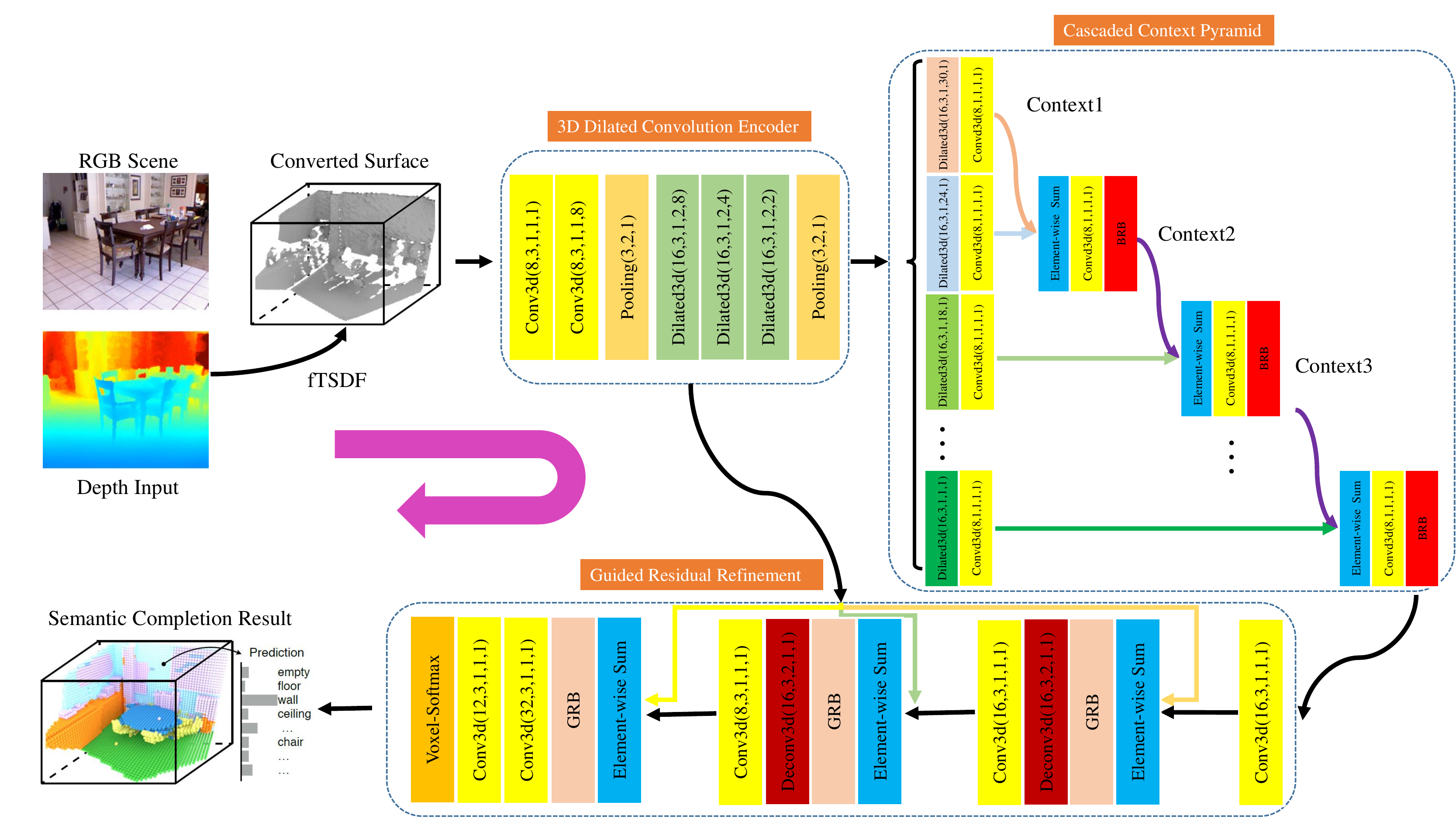} \\
\end{tabular}
}
\caption{Illustration of our Cascaded Context Pyramid Network (CCPNet).
Taking a single-view depth map as input, the CCPNet predicts the occupancy and object labels for each voxel in the view frustum.
%
%
%
With light-weight operations, the CCPNet can produce full-resolution 3D completion results.
The convolution parameters are shown as (number of filters, kernel size, stride, dilation, number of subvolumes).}
\label{fig:framework}
\vspace{-2mm}
\end{figure*}

Although effective, current methods only consider the global semantics, which usually result in low-resolution predictions and lose the scene details.
Different from previous works, we propose to integrate both local geometric details and multi-scale 3D contexts of the scene for the SSC task.
To reduce the semantic gaps of multi-scale 3D contexts, we propose a self-cascaded context aggregation method to generate coherent labeling results.
Meanwhile, the local geometric details are also incorporated to identify the fine-structured objects in a coarse-to-fine manner.
We note that the proposed modules are general-purpose for 3D CNNs.
Thus, they can be easily applied to other 3D tasks.
\section{Cascaded Context Pyramid Network}
Fig.~\ref{fig:framework} illustrates the overall architecture of our CCPNet.
Given a single-view depth map of a 3D scene, the goal of our CCPNet is to map the voxels in the view frustum to one of the semantic labels $C=[c_{0},c_{1},...,c_{N+1}]$, where $N$ is number of semantic categories and $c_{0}$ stands for empty voxels.
Our CCPNet is a self-cascaded pyramid structure to successively aggregate multi-scale 3D contexts and local geometry details for full-resolution scene completions.
It consists of three key components, \emph{i.e.}, 3D Dilated Convolution Encoder (DCE),
Cascaded Context Pyramid (CCP), and Guided Residual Refinement (GRR).
Functionally, the DCE adopts multiple dilated convolutions with separated kernels to extract 3D feature representations from single-view depth images.
Then, the CCP performs the sequential global-to-local context aggregation to improve the labeling coherence.
After the context aggregation, the GRR is introduced to refine the target objects using low-level features learned by the shallow layers.
In the following subsections, we will describe these components in detail.
\subsection{3D Dilated Convolution Encoder}
\textbf{Input Tensor Generation.}
For the input of our front-end 3D DCE, we follow previous works~\cite{song2017semantic,garbade2018two,guo2018view} and rotate the 3D scene to align with the gravity and room orientation based on the Manhattan assumption.
We consider the absolute dimensions of the 3D space with 4.8 m horizontally, 2.88 m vertically, and 4.8 m in depth.
Each 3D scene is encoded into a flipped Truncated Signed Distance Function (fTSDF)~\cite{song2017semantic} with grid size 0.02 m, truncation value 0.24 m, resulting in a $240\times144\times240$ tensor as the network input.
Our method produces the completion result with the same resolution as input.
However, due to the fully convolutional structure and the light-weight network design, our method certainly can take larger depth images as input, even full-resolution depth maps (e.g., 427$\times$561 from depth sensors).
During the model training, we render depth maps from virtual viewpoints of 3D scenes and voxelize the full 3D scenes with object labels as ground truth.

\begin{figure}
\centering
\resizebox{0.5\textwidth}{!}
{
\begin{tabular}{@{}c@{}c@{}}
\includegraphics[width=1\linewidth,height=2.2cm]{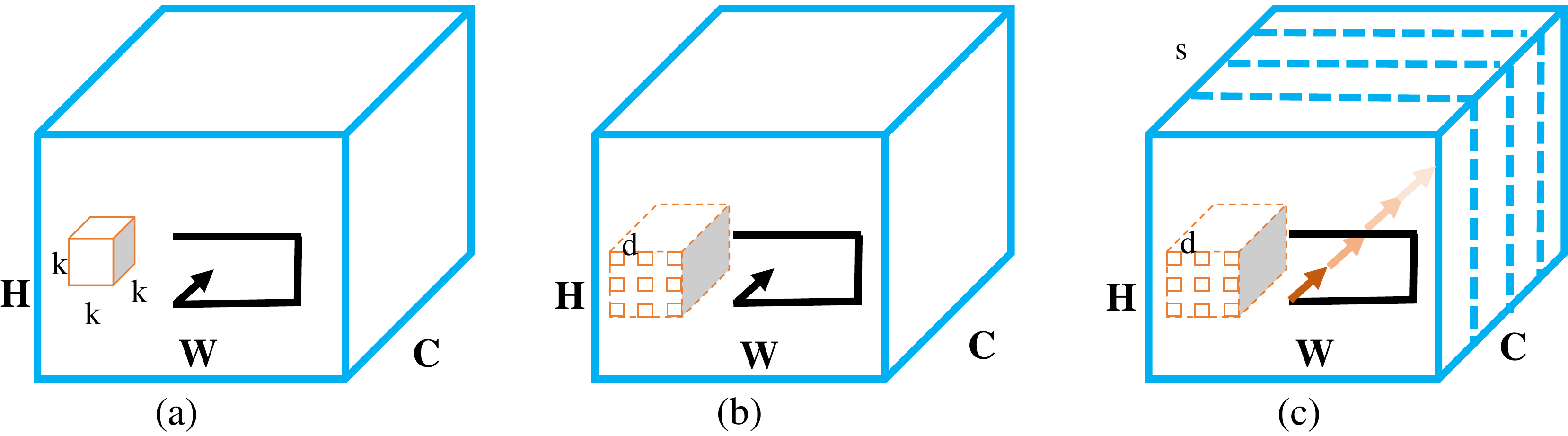} \\
\end{tabular}
}
\caption{Comparison of (a) Vanilla 3D convolution~\cite{ji20133d}, (b) 3D dilated convolution~\cite{song2017semantic} and (c) Our proposed method.}
\vspace{-4mm}
\label{fig:dce}
\end{figure}

\textbf{Encoder Structure.}
Processing 3D data needs large memories and huge computations.
To reduce the memory-requirement, we propose a light-weight encoder to extract the 3D feature representations of scenes, as shown in Fig.~\ref{fig:framework}.
As demonstrated in dense labeling tasks~\cite{zhao2017pyramid,zhang2018agile,chen2018deeplab,zhang2019deep}, large contexts can provide valuable information for understanding the scenes.
For the 3D scenes and depth images, spatial context is more useful due to the lack of high frequency signals.
To effectively learn spatial contextual information, we make sure our encoder has a big enough receptive field.
A direct method is using the 3D dilated convolution proposed in~\cite{yu2016multi,song2017semantic}, which can exponentially expand the receptive field without a loss of resolution or coverage.
%
However, the computation of 3D dilated convolutions is rather huge, because we need to perform convolutions with large volumes.
To address this problem, we propose the 3D dilated convolutions with \emph{separated kernels}.
More specifically, we first separate the input tensor into several subvolumes, then apply the 3D dilated kernels to each subvolume for the convolutions.
The reasons are two-fold. On the one hand, our method can reduce the model parameters and computations, and inherit all characteristics of dilated convolutions.
On the other hand, our method considers the characteristic of depth profiles, in which the depth values are continuous only in neighbour regions.
Fig.~\ref{fig:dce} shows the differences of vanilla 3D convolution~\cite{ji20133d}, 3D dilated convolution~\cite{song2017semantic} and our proposed method.
To build our 3D DCE, we stack the proposed 3D dilated convolution several times with 3D pooling.
Besides, to avoid the extreme separation, we reduce the number of subvolumes along with the network depth.
The detailed parameters are shown in Fig.~\ref{fig:framework}.
\vspace{-2mm}
\subsection{Cascaded Context Pyramid}
\begin{figure}
\centering
\resizebox{0.5\textwidth}{!}
{
\begin{tabular}{@{}c@{}c@{}}
\includegraphics[width=1\linewidth,height=3.4cm]{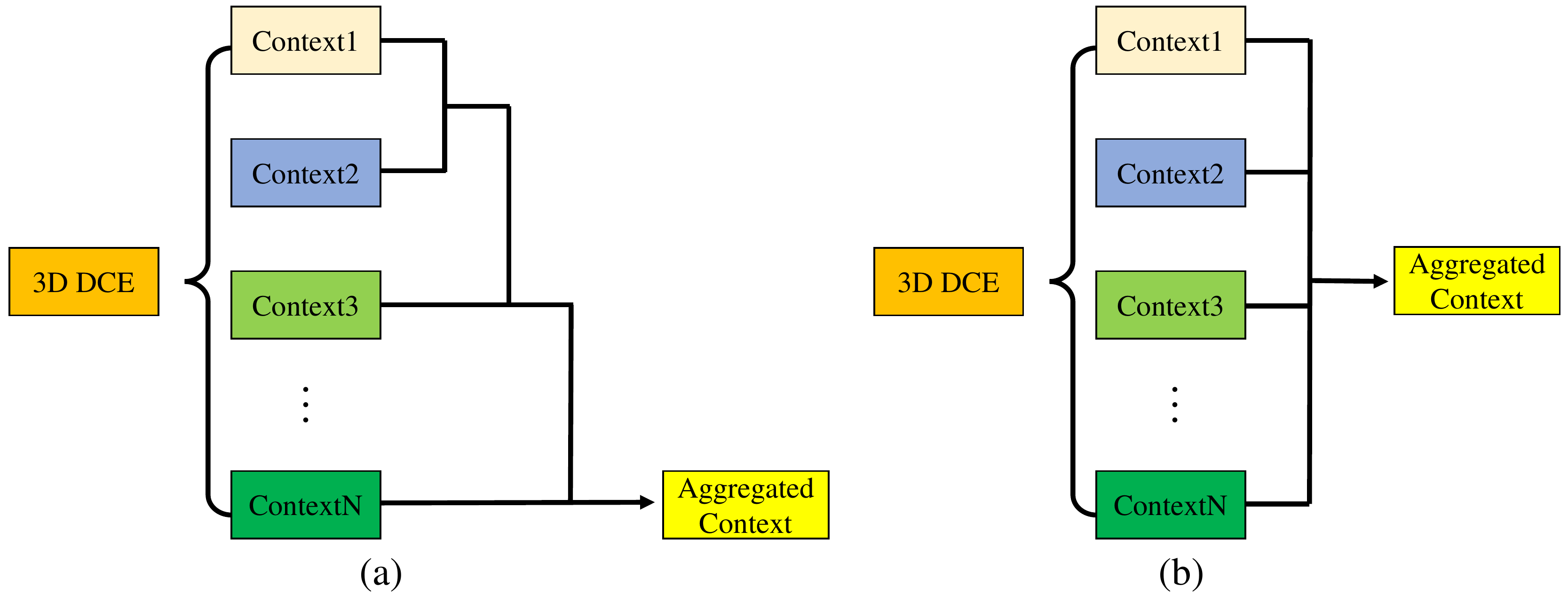} \\
\end{tabular}
}
\caption{Comparison of different multi-scale context aggregation methods. (a) Our self-cascaded context aggregation approach, which reduces the semantic gaps of different scales. (b) Existing parallel concatenations, such as PSPNet~\cite{zhao2017pyramid}, Deeplab variants~\cite{chen2018deeplab}. ``Context'' denotes the dilated convolution for context extraction.}
\vspace{-2mm}
\label{fig:ccp}
\end{figure}

For scene completion, different objects have very different physical 3D sizes and visual orientations.
This implies that the model needs to capture information at different contexts in order to recognize objects reliably.
Besides, for confusing manmade objects in indoor scenes, obtaining coherent labeling results is not easily accessible,
because they are of high intra-class variance and low inter-class variance.
Therefore, it is insufficient to use only the single-scale and global information of the target objects~\cite{song2017semantic,liu2018semantic}.
We need to introduce multi-scale context information, which characterizes the underlying dependencies between an
object and its surroundings.
However, it is very hard to retain the hierarchical dependencies in contexts of different scales, using common fusion strategies (\emph{e.g.}, direct stack~\cite{chen2018deeplab,zhao2017pyramid}).
To address this issue, we propose a novel self-cascaded context pyramid architecture, as shown in Fig.~\ref{fig:ccp} (a).
Different from previous methods, our method sequentially aggregates the global-to-local contexts while well retains the hierarchical dependencies, \emph{i.e.}, the underlying inclusion and location relationship among the objects and scenes in different scales.

\textbf{Architecture Details.} To build the context pyramid, we perform 3D dilated convolutions on the last pooling layer of the 3D DCE to capture multi-scale contexts.
By setting varied dilation rates (30, 24, 18, 12, 6 and 1 in the experiments) and feature reduction layers, a series of
3D feature maps with global-to-local contexts are generated.
The large-scale context contains more semantics and wider visual cues, while the small-scale context retains object geometry details.
Meanwhile, the obtained feature maps with multi-scale contexts can be aligned automatically due to their equal resolution.
To well retain the hierarchical dependencies of multi-scale contexts, we \emph{sequentially} aggregate them in a self-cascaded pyramid manner.
Formally, it can be described as:
\begin{equation}
\label{equ:equ1}
X_{sa}=
\left\{
\begin{aligned}
&f(\cdots f(f(X_{1}\oplus X_{2})\oplus X_{3})\oplus\cdots\oplus X_{n}),\\
&d_{1}>d_{2}>d_{3}>\cdots>d_{n}.
\end{aligned}
\right.
\end{equation}
where $X_{n}$ denotes the $n$-scale context, $X_{sa}$ is the final aggregated context and $d_{n}$ is the dilation rate for extracting the context $X_{n}$.
$\oplus$ denotes the element-wise summation. $f$ denotes the Basic Residual Block (BRB)~\cite{he2016deep}, as shown in Fig.~\ref{fig:grb} (a).
In our proposed method, we first aggregate the large-scale context with big dilation rates, then the context with small dilation rates.
This aggregation rule is consistent with the human visual mechanism, \emph{i.e.}, large-scale context could play a guiding role in integrating small-scale context.

We also notice that there are other outstanding structures for multi-scale contexts, such as PPM~\cite{zhao2017pyramid} and ASPP~\cite{chen2018deeplab}, as shown in Fig.~\ref{fig:ccp} (b).
In order to aggregate information with different contexts, they add a layer that \emph{parallelly} concatenates the feature maps with different receptive fields:
\begin{equation}
\label{equ:equ2}
X_{pa}=
\left\{
\begin{aligned}
&g([X_{1},X_{2},X_{3},\cdots,X_{n}]),\\
&d_{1}>d_{2}>d_{3}>\cdots>d_{n}.
\end{aligned}
\right.
\end{equation}
where $g$ denotes the aggregation function, which usually is an $1\times1\times1$ convolutional layer.
$\lbrack\cdots\rbrack$ is the concatenation operation in channel-wise.
However, our proposed self-cascaded pyramid architecture has several advantages:
1) Our self-cascaded strategy enhances the hierarchical dependencies in different context scales. Thus, it is more effective than the parallel strategies such as PSPNet~\cite{zhao2017pyramid}, DeepLab variants~\cite{chen2018deeplab},
which directly fuse the multi-scale contexts with large semantic gaps;
2) Our method introduces more complicated nonlinear operations (Equ. 1), thus it has a stronger capacity to model the relationship of different contexts than simple convolution operations.
3) By adopting the summation, the sequential aggregation significantly reduces the parameters and computations.
Experiments also verify the effectiveness of our proposed method.
\subsection{Guided Residual Refinement}
\begin{figure}
\centering
\resizebox{0.48\textwidth}{!}
{
\begin{tabular}{@{}c@{}c@{}}
\includegraphics[width=1\linewidth,height=5cm]{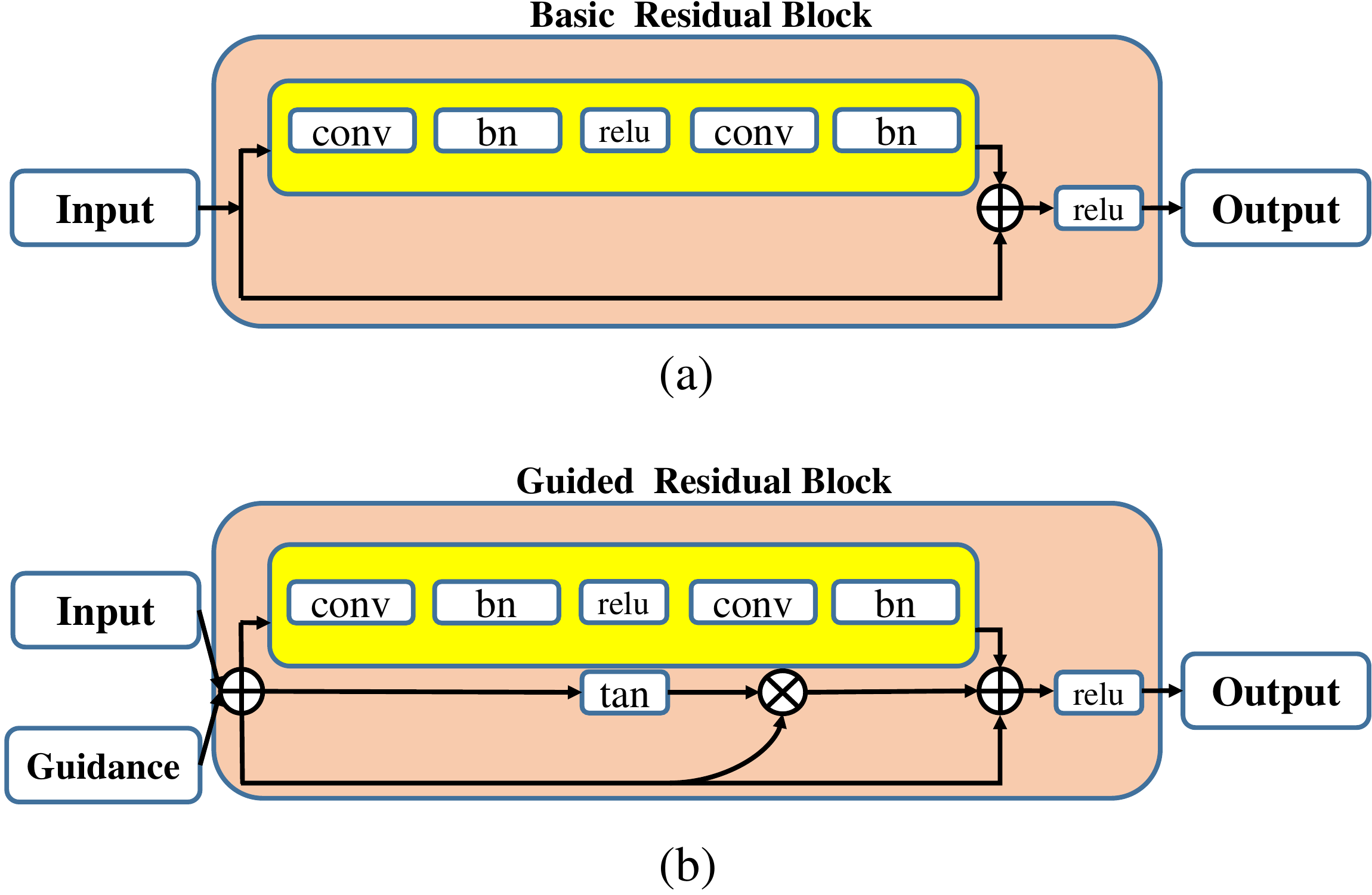} \\
\end{tabular}
}
\caption{The used residual modules in our CCPNet. (a) The Basic Residual Block (BRB)~\cite{he2016deep}. (b) The proposed Guided Residual Block (GRB). In the GRB, we add a tangent function-based connection to amplify the fused features.}
\vspace{-4mm}
\label{fig:grb}
\end{figure}
Besides semantic confusing categories, fine-structured objects also increase the difficulty for
accurate labeling in 3D scenes.
However, current methods usually produce low-resolution predictions, thus it is very hard to retain the fine-grained details of objects.
To address this problem, we propose to reuse low-level features with the Guided Residual Refinement (GRR), as shown in the bottom of Fig.~\ref{fig:framework}.
Specifically, the rich low-level features are progressively reintroduced into the prediction stream by guided residual connections.
As a result, the coarse feature maps can be refined and the low-level details can be restored for full-resolution predictions.
The used Guided Residual Block (GRB) is shown in Fig.~\ref{fig:grb} (b), which can be formulated as:
\begin{align}
\hat{X}&=X\oplus G,\\
X_{rf}&=ReLu(\hat{X}\oplus\hat{X}Tanh(\hat{X})\oplus h(\hat{X}))\\
&= ReLu(\hat{X}(I\oplus Tanh(\hat{X}))\oplus h(\hat{X}))\\
&= ReLu(\hat{X}_{G}\oplus h(\hat{X})).
\label{equ:equ3}
\vspace{-2mm}
\end{align}
where $X$ is the input semantic context feature and $G$ is the guidance feature coming from a shallower layer.
$\oplus$ denotes the element-wise summation and $h$ is the standard non-linear transform in residual blocks.
$X_{rf}$ is the refined feature map.
$ReLu(\cdot)$ and $Tanh(\cdot)$ are the rectified linear unit and hyperbolic tangent activation, respectively.
To restore finer details with the shallower layer, we first integrate the input feature and the guidance (Equ. 3), then we introduce an auxiliary connection to the BRB~\cite{he2016deep}.
More specifically, we use the hyperbolic tangent activation to amplify the integrated features (resulting in $\hat{X}_{G}$), as shown in Fig.~\ref{fig:grb} (b) and Equ. 4-6.
It is very beneficial to fuse low-level features by the guided refinement strategy.
On the one hand, the feature maps of $X$ and $G$ represent different semantics at varied levels.
Thus, due to their inherent semantic gaps, directly stacking all these features~\cite{hariharan2015hypercolumns,ronneberger2015u,cciccek20163d} may not be an efficient strategy.
In the proposed method, the influence of semantic gaps is alleviated when a residual iteration strategy is adopted~\cite{greff2016highway}.
On the other hand, the feature amplification connection enhances the effect of low-level details and gradient propagations, which helps the effectively end-to-end training.
There also exist effective refinement strategies for detail enhancement~\cite{pinheiro2016learning,lin2017refinenet,zhang2017amulet}.
However, they are very different from ours.
First, our strategy focuses on amplifying low-level features considering the 3D data properties, \emph{e.g.}, high computation and memory requirements.
In contrast, previous methods introduce complex refinement modules, which are hardly executable for the 3D data.
Besides, we only choose specific shallow layers for the refinement, as shown in the bottom of Fig.~\ref{fig:framework}.
Other methods incorporate all the hierarchical layers that inevitably contain boundary noises~\cite{pinheiro2016learning,lin2017refinenet}.
To build our model, several GRB modules are elaborately embedded in the prediction part, which can greatly prevent the fitting residual from accumulating.
As a result, the proposed CCPNet effectively works in a coarse-to-fine labeling manner for full-resolution predictions.
\subsection{Network Training}
Given the training dataset (\emph{i.e.}, the paired depth images and ground truth volumetric labels of 3D scenes), our proposed CCPNet can be trained in an end-to-end manner.
We adopt the voxel-wise softmax loss function~\cite{song2017semantic} for the network training.
The loss can be expressed as:
\begin{equation}
L(p,y) =\sum_{i,j,k} w_{ijk}L_{sm}(p_{ijk},y_{ijk}),
\vspace{-2mm}
\end{equation}
where $L_{sm}$ is the softmax cross-entropy loss, $y_{ijk}$ is the ground truth label, $p_{ijk}$ is the predicted probability of the voxel at coordinates $(i,j,k)$.
The weight $w_{ijk}\in\{0,1\}$ is used to balance the loss between different semantic categories.
Due to the sparsity of 3D data, the ratio of empty vs. occupied voxels is extremely imbalanced.
To address this problem, we follow~\cite{song2017semantic} and randomly sample the training voxels with a 2:1 ratio to ensure that each mini-batch has a balanced set of empty and occupied examples.
\begin{table*}
\begin{center}
\doublerulesep=0.5pt
\resizebox{0.9\textwidth}{!}
{
\begin{tabular}{|c|ccccc|c|c|c|c|c|c|c|c|c|c|c|c|c|c|c|c|c|c|c|c|c|c|c|c|c|c|c|c|c|c|c|c|c|c|c|c|c|c|c|c|c|c|c|c|c|c|c|c|c|c|c|c|c|c|c|c|c|c|c|}
\hline
\multicolumn{4}{|c|}{}&\multicolumn{6}{|c|}{scene completion}&\multicolumn{24}{|c|}{semantic scene completion}
\\
\hline
\multicolumn{4}{|c|}{Methods}
&\multicolumn{2}{|c}{prec.}&\multicolumn{2}{c}{recall}&\multicolumn{2}{c|}{IoU}
&\multicolumn{2}{|c}{ceil.}&\multicolumn{2}{c}{floor}&\multicolumn{2}{c}{wall}
&\multicolumn{2}{c}{win.}&\multicolumn{2}{c}{chair}&\multicolumn{2}{c}{bed}
&\multicolumn{2}{c}{sofa}&\multicolumn{2}{c}{table}&\multicolumn{2}{c}{tvs}
&\multicolumn{2}{c}{furn.}&\multicolumn{2}{c}{objs.}&\multicolumn{2}{|c|}{avg.}
\\
\hline
\multicolumn{4}{|c|}{SSCNet~\cite{song2017semantic}}
&\multicolumn{2}{|c}{76.3}&\multicolumn{2}{c}{95.2}&\multicolumn{2}{c|}{73.5}
&\multicolumn{2}{|c}{96.3}&\multicolumn{2}{c}{84.9}&\multicolumn{2}{c}{56.8}
&\multicolumn{2}{c}{28.2}&\multicolumn{2}{c}{21.3}&\multicolumn{2}{c}{56.0}
&\multicolumn{2}{c}{52.7}&\multicolumn{2}{c}{33.7}&\multicolumn{2}{c}{10.9}
&\multicolumn{2}{c}{44.3}&\multicolumn{2}{c}{25.4}&\multicolumn{2}{|c|}{46.4}
\\
\multicolumn{4}{|c|}{VVNet~\cite{guo2018view}}
&\multicolumn{2}{|c}{90.8}&\multicolumn{2}{c}{91.7}&\multicolumn{2}{c|}{84.0}
&\multicolumn{2}{|c}{98.4}&\multicolumn{2}{c}{87.0}&\multicolumn{2}{c}{61.0}
&\multicolumn{2}{c}{54.8}&\multicolumn{2}{c}{49.3}&\multicolumn{2}{c}{83.0}
&\multicolumn{2}{c}{75.5}&\multicolumn{2}{c}{55.1}&\multicolumn{2}{c}{43.5}
&\multicolumn{2}{c}{68.8}&\multicolumn{2}{c}{57.7}&\multicolumn{2}{|c|}{66.7}
\\
\multicolumn{4}{|c|}{DCRF~\cite{zhang2018semantic}}
&\multicolumn{2}{|c}{--}&\multicolumn{2}{c}{--}&\multicolumn{2}{c|}{--}
&\multicolumn{2}{|c}{95.4}&\multicolumn{2}{c}{84.3}&\multicolumn{2}{c}{57.7}
&\multicolumn{2}{c}{24.5}&\multicolumn{2}{c}{28.2}&\multicolumn{2}{c}{63.4}
&\multicolumn{2}{c}{55.3}&\multicolumn{2}{c}{34.5}&\multicolumn{2}{c}{19.6}
&\multicolumn{2}{c}{45.8}&\multicolumn{2}{c}{28.7}&\multicolumn{2}{|c|}{48.8}
\\
\multicolumn{4}{|c|}{ESSCNet~\cite{zhang2018efficient}}
&\multicolumn{2}{|c}{92.6}&\multicolumn{2}{c}{90.4}&\multicolumn{2}{c|}{84.5}
&\multicolumn{2}{|c}{96.6}&\multicolumn{2}{c}{83.7}&\multicolumn{2}{c}{74.9}
&\multicolumn{2}{c}{59.0}&\multicolumn{2}{c}{55.1}&\multicolumn{2}{c}{83.3}
&\multicolumn{2}{c}{78.0}&\multicolumn{2}{c}{61.5}&\multicolumn{2}{c}{47.4}
&\multicolumn{2}{c}{73.5}&\multicolumn{2}{c}{62.9}&\multicolumn{2}{|c|}{70.5}
\\
\multicolumn{4}{|c|}{SATNet~\cite{liu2018see}}
&\multicolumn{2}{|c}{80.7}&\multicolumn{2}{c}{96.5}&\multicolumn{2}{c|}{78.5}
&\multicolumn{2}{c}{97.9}&\multicolumn{2}{c}{82.5}&\multicolumn{2}{c}{57.7}
&\multicolumn{2}{c}{58.5}&\multicolumn{2}{c}{45.1}&\multicolumn{2}{c}{78.4}
&\multicolumn{2}{c}{72.3}&\multicolumn{2}{c}{47.3}&\multicolumn{2}{c}{45.7}
&\multicolumn{2}{c}{67.1}&\multicolumn{2}{c}{55.2}&\multicolumn{2}{|c|}{64.3}
\\
\multicolumn{4}{|c|}{Ours}
&\multicolumn{2}{|c}{\textbf{98.2}}&\multicolumn{2}{c}{\textbf{96.8}}&\multicolumn{2}{c|}{\textbf{91.4}}
&\multicolumn{2}{|c}{\textbf{99.2}}&\multicolumn{2}{c}{\textbf{89.3}}&\multicolumn{2}{c}{\textbf{76.2}}
&\multicolumn{2}{c}{\textbf{63.3}}&\multicolumn{2}{c}{\textbf{58.2}}&\multicolumn{2}{c}{\textbf{86.1}}
&\multicolumn{2}{c}{\textbf{82.6}}&\multicolumn{2}{c}{\textbf{65.6}}&\multicolumn{2}{c}{\textbf{53.2}}
&\multicolumn{2}{c}{\textbf{76.8}}&\multicolumn{2}{c}{\textbf{65.2}}&\multicolumn{2}{|c|}{\textbf{74.2}}
\\
\hline
\end{tabular}
}
\caption{The performances of different scene completion methods on the SUNCG dataset. The best results are in bold.}
\label{table:SUNCG}
\vspace{-6mm}
\end{center}
\end{table*}
\begin{figure*}
\centering
\resizebox{1\textwidth}{!}
{
\begin{tabular}{@{}c@{}c@{}c@{}c@{}c@{}c@{}c@{}c@{}c@{}c@{}c@{}c@{}c@{}c@{}c@{}c@{}c@{}c@{}c}
(a)\hspace{2.2cm}(b)\hspace{2cm}(c)\hspace{2cm}(d)\hspace{2cm}(e)\hspace{2cm}(f)\hspace{2.2cm}(g)\\
\includegraphics[width=1\linewidth,height=6cm]{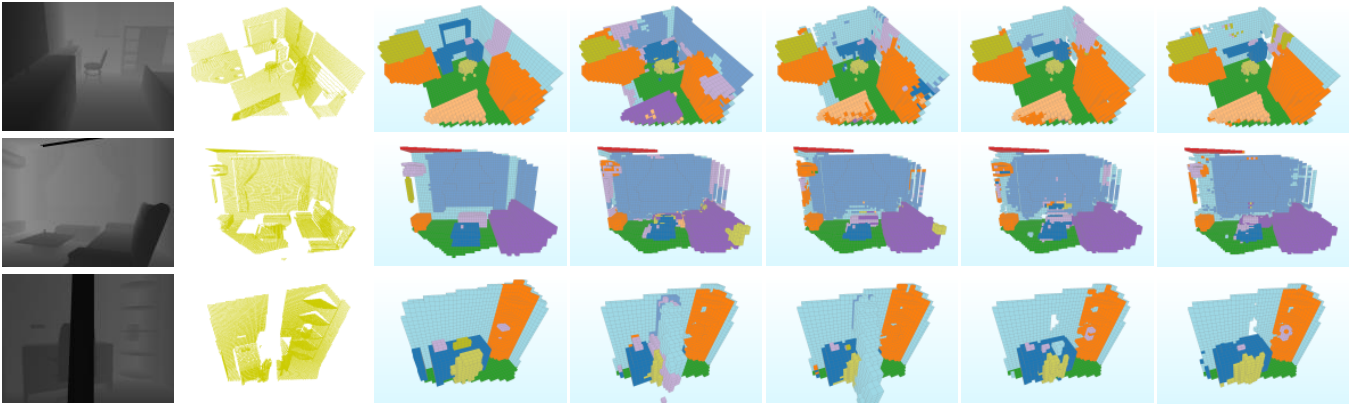} \\
\includegraphics[width=0.8\linewidth,height=0.5cm]{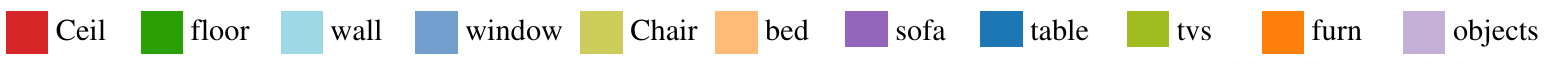} \\
\end{tabular}
}
\caption{Completion results with different methods on the SUNCG dataset.
From the left to right: (a) Input Depth; (b) fTSDF Surface; (c) Ground Truth; (d) SSCNet~\cite{song2017semantic}; (e) VVNet~\cite{guo2018view}; (f) ESSCNet~\cite{zhang2018efficient}; (g) Ours.
It can be observed that, our results constantly contain more accurate and detailed structures compared to the baselines.
The figure is best viewed in color with 200\% zooming-in.}
\label{fig:SUNCG}
\vspace{-4mm}
\end{figure*}
\section{Experiments}
\subsection{Experimental Settings}
\textbf{Datasets.}
The synthetic \textbf{SUNCG} dataset~\cite{song2016deep} consists of 45622 indoor scenes.
Technically, the depth images and semantic scene volumes can be acquired by setting different camera orientations.
Following previous useful works~\cite{song2017semantic,garbade2018two,guo2018view}, we adopt the same training/test split for our network training and evaluation.
More specifically, the training set contains about 150K depth images and the corresponding ground truth volumes.
The test set consists of totally 470 pairs sampled from 170 non-overlap scenes.

The real \textbf{NYU} dataset~\cite{silberman2012indoor} includes 1449 depth images captured by the Kinect depth sensor.
The depth images are partitioned into 795 for training and 654 for test.
Following previous works, we adopt the ground truth completion and segmentation from~\cite{guo2015predicting}.
Some labeled volumes and their corresponding depth images are not well aligned in the NYU dataset.
Thus, we also use the NYU CAD dataset~\cite{firman2016structured}, in which the depth map is rendered from the label volume.
The NYU dataset is challenging due to the unavoidably measurement errors in the depth images collected by Kinect.
As in~\cite{song2017semantic,guo2018view,liu2018see}, we also pre-train the network on the SUNCG dataset before fine-tuning it on the NYU dataset.

\textbf{Evaluation Metrics.} We mainly follow~\cite{song2017semantic} and report the precision, recall, and Intersection over Union (IoU) of the compared methods.
The IoU measures the overlapped ratio between intersection and union of the positive prediction volume and
the ground truth volume.
In this work, two tasks are considered: Scene Completion (SC) and Semantic Scene Completion (SSC).
For the SC task, we treat all voxels as binary predictions, \emph{i.e.}, occupied or non-occupied.
The ground truth volume includes all the occluded voxels in the view frustum.
For the SSC task, we report the IoU of each class, and average them to get the mean IoU.

\textbf{Training Protocol.} We implement our CCPNet in the modified Caffe toolbox~\cite{jia2014caffe} for 3D data processing.
We perform experiments on a quad-core PC with an Intel i4790 CPU and one NVIDIA TITAN X GPU (12G memory).
For the CCPNet, we initialize the weights by the ``msra'' method~\cite{he2015delving}.
During the training, we use the standard SGD method with a batch size 4, momentum 0.9 and weight decay 0.0005.
We set the base learning rate to 0.01.
For the SUNCG dataset, we train the CCPNet with 200K iterations and change the learning rate to 0.001 after 150K iterations.
To reduce the performance bias, we evaluate the results at every 5K steps after 180K iterations, and average them as the final results.
For both the NYU Kinect and NYU CAD datasets, we follow previous works~\cite{song2017semantic,guo2018view,garbade2018two,zhang2018efficient,liu2018see}, and fine-tune the CPPNet pre-trained from the SUNCG dataset with 10K iterations.
After that, we test the models at every 2K iterations and pick the best one as the final result.
\vspace{-2mm}
\subsection{Experimental Results}
\begin{table*}
\begin{center}
\doublerulesep=0.5pt
\resizebox{0.9\textwidth}{!}
{
\begin{tabular}{|c|ccccc|c|c|c|c|c|c|c|c|c|c|c|c|c|c|c|c|c|c|c|c|c|c|c|c|c|c|c|c|c|c|c|c|c|c|c|c|c|c|c|c|c|c|c|c|c|c|c|c|c|c|c|c|c|c|c|c|c|c|c|}
\hline
\multicolumn{8}{|c|}{}&\multicolumn{6}{|c|}{scene completion}&\multicolumn{24}{|c|}{semantic scene completion}
\\
\hline
\multicolumn{4}{|c|}{Methods}&\multicolumn{4}{|c|}{Trained on}
&\multicolumn{2}{|c}{prec.}&\multicolumn{2}{c}{recall}&\multicolumn{2}{c|}{IoU}
&\multicolumn{2}{|c}{ceil.}&\multicolumn{2}{c}{floor}&\multicolumn{2}{c}{wall}
&\multicolumn{2}{c}{win.}&\multicolumn{2}{c}{chair}&\multicolumn{2}{c}{bed}
&\multicolumn{2}{c}{sofa}&\multicolumn{2}{c}{table}&\multicolumn{2}{c}{tvs}
&\multicolumn{2}{c}{furn.}&\multicolumn{2}{c}{objs.}&\multicolumn{2}{|c|}{avg.}
\\
\hline
\multicolumn{4}{|c|}{Lin et al.~\cite{lin2013holistic}}&\multicolumn{4}{|c|}{NYU}
&\multicolumn{2}{|c}{58.5}&\multicolumn{2}{c}{49.9}&\multicolumn{2}{c|}{36.4}
&\multicolumn{2}{|c}{0.0}&\multicolumn{2}{c}{11.7}&\multicolumn{2}{c}{13.3}
&\multicolumn{2}{c}{14.1}&\multicolumn{2}{c}{9.4}&\multicolumn{2}{c}{29.0}
&\multicolumn{2}{c}{24.0}&\multicolumn{2}{c}{6.0}&\multicolumn{2}{c}{7.0}
&\multicolumn{2}{c}{16.2}&\multicolumn{2}{c}{1.1}&\multicolumn{2}{|c|}{12.0}
\\
\multicolumn{4}{|c|}{Geiger and Wang~\cite{geiger2015joint}}&\multicolumn{4}{|c|}{NYU}
&\multicolumn{2}{|c}{65.7}&\multicolumn{2}{c}{58.0}&\multicolumn{2}{c|}{44.4}
&\multicolumn{2}{|c}{10.2}&\multicolumn{2}{c}{62.5}&\multicolumn{2}{c}{19.1}
&\multicolumn{2}{c}{5.8}&\multicolumn{2}{c}{8.5}&\multicolumn{2}{c}{40.6}
&\multicolumn{2}{c}{27.7}&\multicolumn{2}{c}{7.0}&\multicolumn{2}{c}{6.0}
&\multicolumn{2}{c}{22.6}&\multicolumn{2}{c}{5.9}&\multicolumn{2}{|c|}{19.6}
\\
\hline
\multicolumn{4}{|c|}{SSCNet~\cite{song2017semantic}}&\multicolumn{4}{|c|}{NYU}
&\multicolumn{2}{|c}{57.0}&\multicolumn{2}{c}{94.5}&\multicolumn{2}{c|}{55.1}
&\multicolumn{2}{|c}{15.1}&\multicolumn{2}{c}{94.7}&\multicolumn{2}{c}{24.4}
&\multicolumn{2}{c}{0.0}&\multicolumn{2}{c}{12.6}&\multicolumn{2}{c}{32.1}
&\multicolumn{2}{c}{35.0}&\multicolumn{2}{c}{13.0}&\multicolumn{2}{c}{7.8}
&\multicolumn{2}{c}{27.1}&\multicolumn{2}{c}{10.1}&\multicolumn{2}{|c|}{24.7}
\\
\multicolumn{4}{|c|}{SSCNet~\cite{song2017semantic}}&\multicolumn{4}{|c|}{SUNCG}
&\multicolumn{2}{|c}{55.6}&\multicolumn{2}{c}{91.9}&\multicolumn{2}{c|}{53.2}
&\multicolumn{2}{|c}{5.8}&\multicolumn{2}{c}{81.8}&\multicolumn{2}{c}{19.6}
&\multicolumn{2}{c}{5.4}&\multicolumn{2}{c}{12.9}&\multicolumn{2}{c}{34.4}
&\multicolumn{2}{c}{26.0}&\multicolumn{2}{c}{13.6}&\multicolumn{2}{c}{6.1}
&\multicolumn{2}{c}{9.4}&\multicolumn{2}{c}{7.4}&\multicolumn{2}{|c|}{20.2}
\\
\multicolumn{4}{|c|}{SSCNet~\cite{song2017semantic}}&\multicolumn{4}{|c|}{NYU+SUNCG}
&\multicolumn{2}{|c}{59.3}&\multicolumn{2}{c}{92.9}&\multicolumn{2}{c|}{56.6}
&\multicolumn{2}{|c}{15.1}&\multicolumn{2}{c}{94.6}&\multicolumn{2}{c}{24.7}
&\multicolumn{2}{c}{10.8}&\multicolumn{2}{c}{17.3}&\multicolumn{2}{c}{53.2}
&\multicolumn{2}{c}{45.9}&\multicolumn{2}{c}{15.9}&\multicolumn{2}{c}{13.9}
&\multicolumn{2}{c}{31.1}&\multicolumn{2}{c}{12.6}&\multicolumn{2}{|c|}{30.5}
\\
\multicolumn{4}{|c|}{CSSCNet~\cite{guedes2018semantic}}&\multicolumn{4}{|c|}{NYU+SUNCG}
&\multicolumn{2}{|c}{62.5}&\multicolumn{2}{c}{82.3}&\multicolumn{2}{c|}{54.3}
&\multicolumn{2}{|c}{--}&\multicolumn{2}{c}{--}&\multicolumn{2}{c}{--}
&\multicolumn{2}{c}{--}&\multicolumn{2}{c}{--}&\multicolumn{2}{c}{--}
&\multicolumn{2}{c}{--}&\multicolumn{2}{c}{--}&\multicolumn{2}{c}{--}
&\multicolumn{2}{c}{--}&\multicolumn{2}{c}{--}&\multicolumn{2}{|c|}{27.5}
\\
\multicolumn{4}{|c|}{VVNet~\cite{guo2018view}}&\multicolumn{4}{|c|}{NYU+SUNCG}
&\multicolumn{2}{|c}{69.8}&\multicolumn{2}{c}{83.1}&\multicolumn{2}{c|}{61.1}
&\multicolumn{2}{|c}{19.3}&\multicolumn{2}{c}{94.8}&\multicolumn{2}{c}{28.0}
&\multicolumn{2}{c}{12.2}&\multicolumn{2}{c}{19.6}&\multicolumn{2}{c}{57.0}
&\multicolumn{2}{c}{50.5}&\multicolumn{2}{c}{17.6}&\multicolumn{2}{c}{11.9}
&\multicolumn{2}{c}{35.6}&\multicolumn{2}{c}{15.3}&\multicolumn{2}{|c|}{32.9}
\\
\multicolumn{4}{|c|}{DCRF~\cite{zhang2018semantic}}&\multicolumn{4}{|c|}{NYU}
&\multicolumn{2}{|c}{--}&\multicolumn{2}{c}{--}&\multicolumn{2}{c|}{--}
&\multicolumn{2}{|c}{18.1}&\multicolumn{2}{c}{92.6}&\multicolumn{2}{c}{27.1}
&\multicolumn{2}{c}{10.8}&\multicolumn{2}{c}{18.8}&\multicolumn{2}{c}{54.3}
&\multicolumn{2}{c}{47.9}&\multicolumn{2}{c}{17.1}&\multicolumn{2}{c}{15.1}
&\multicolumn{2}{c}{34.7}&\multicolumn{2}{c}{13.0}&\multicolumn{2}{|c|}{31.8}
\\
\multicolumn{4}{|c|}{TS3D,V2~\cite{garbade2018two}}&\multicolumn{4}{|c|}{NYU}
&\multicolumn{2}{|c}{65.7}&\multicolumn{2}{c}{87.9}&\multicolumn{2}{c|}{60.4}
&\multicolumn{2}{|c}{8.9}&\multicolumn{2}{c}{94.0}&\multicolumn{2}{c}{26.4}
&\multicolumn{2}{c}{16.1}&\multicolumn{2}{c}{14.2}&\multicolumn{2}{c}{53.5}
&\multicolumn{2}{c}{45.8}&\multicolumn{2}{c}{16.4}&\multicolumn{2}{c}{13.0}
&\multicolumn{2}{c}{32.9}&\multicolumn{2}{c}{12.7}&\multicolumn{2}{|c|}{30.4}
\\
\multicolumn{4}{|c|}{TS3D,V3+~\cite{garbade2018two}}&\multicolumn{4}{|c|}{NYU}
&\multicolumn{2}{|c}{64.9}&\multicolumn{2}{c}{88.8}&\multicolumn{2}{c|}{60.2}
&\multicolumn{2}{|c}{8.2}&\multicolumn{2}{c}{94.1}&\multicolumn{2}{c}{26.4}
&\multicolumn{2}{c}{19.2}&\multicolumn{2}{c}{17.2}&\multicolumn{2}{c}{55.5}
&\multicolumn{2}{c}{48.4}&\multicolumn{2}{c}{16.4}&\multicolumn{2}{c}{22.0}
&\multicolumn{2}{c}{34.0}&\multicolumn{2}{c}{17.1}&\multicolumn{2}{|c|}{32.6}
\\
\multicolumn{4}{|c|}{ESSCNet~\cite{zhang2018efficient}}&\multicolumn{4}{|c|}{NYU}
&\multicolumn{2}{|c}{71.9}&\multicolumn{2}{c}{71.9}&\multicolumn{2}{c|}{56.2}
&\multicolumn{2}{|c}{17.5}&\multicolumn{2}{c}{75.4}&\multicolumn{2}{c}{25.8}
&\multicolumn{2}{c}{6.7}&\multicolumn{2}{c}{15.3}&\multicolumn{2}{c}{53.8}
&\multicolumn{2}{c}{42.4}&\multicolumn{2}{c}{11.2}&\multicolumn{2}{c}{0.0}
&\multicolumn{2}{c}{33.4}&\multicolumn{2}{c}{11.8}&\multicolumn{2}{|c|}{26.7}
\\
\multicolumn{4}{|c|}{SATNet~\cite{liu2018see}}&\multicolumn{4}{|c|}{NYU+SUNCG}
&\multicolumn{2}{|c}{67.3}&\multicolumn{2}{c}{85.8}&\multicolumn{2}{c|}{60.6}
&\multicolumn{2}{|c}{17.3}&\multicolumn{2}{c}{92.1}&\multicolumn{2}{c}{28.0}
&\multicolumn{2}{c}{16.6}&\multicolumn{2}{c}{19.3}&\multicolumn{2}{c}{57.5}
&\multicolumn{2}{c}{53.8}&\multicolumn{2}{c}{17.7}&\multicolumn{2}{c}{18.5}
&\multicolumn{2}{c}{38.4}&\multicolumn{2}{c}{18.9}&\multicolumn{2}{|c|}{34.4}
\\
\multicolumn{4}{|c|}{DDRNet~\cite{li2019rgbd}}&\multicolumn{4}{|c|}{NYU}
&\multicolumn{2}{|c}{71.5}&\multicolumn{2}{c}{80.8}&\multicolumn{2}{c|}{61.0}
&\multicolumn{2}{|c}{21.1}&\multicolumn{2}{c}{92.2}&\multicolumn{2}{c}{33.5}
&\multicolumn{2}{c}{6.8}&\multicolumn{2}{c}{14.8}&\multicolumn{2}{c}{48.3}
&\multicolumn{2}{c}{42.3}&\multicolumn{2}{c}{13.2}&\multicolumn{2}{c}{13.9}
&\multicolumn{2}{c}{35.3}&\multicolumn{2}{c}{13.2}&\multicolumn{2}{|c|}{30.4}
\\
\multicolumn{4}{|c|}{Ours}&\multicolumn{4}{|c|}{NYU}
&\multicolumn{2}{|c}{74.2}&\multicolumn{2}{c}{90.8}&\multicolumn{2}{c|}{63.5}
&\multicolumn{2}{|c}{23.5}&\multicolumn{2}{c}{96.3}&\multicolumn{2}{c}{35.7}
&\multicolumn{2}{c}{20.2}&\multicolumn{2}{c}{25.8}&\multicolumn{2}{c}{61.4}
&\multicolumn{2}{c}{56.1}&\multicolumn{2}{c}{18.1}&\multicolumn{2}{c}{28.1}
&\multicolumn{2}{c}{37.8}&\multicolumn{2}{c}{20.1}&\multicolumn{2}{|c|}{38.5}
\\
\multicolumn{4}{|c|}{Ours}&\multicolumn{4}{|c|}{NYU+SUNCG}
&\multicolumn{2}{|c}{\textbf{78.8}}&\multicolumn{2}{c}{\textbf{94.3}}&\multicolumn{2}{c|}{\textbf{67.1}}
&\multicolumn{2}{|c}{\textbf{25.5}}&\multicolumn{2}{c}{\textbf{98.5}}&\multicolumn{2}{c}{\textbf{38.8}}
&\multicolumn{2}{c}{\textbf{27.1}}&\multicolumn{2}{c}{\textbf{27.3}}&\multicolumn{2}{c}{\textbf{64.8}}
&\multicolumn{2}{c}{\textbf{58.4}}&\multicolumn{2}{c}{\textbf{21.5}}&\multicolumn{2}{c}{\textbf{30.1}}
&\multicolumn{2}{c}{\textbf{38.4}}&\multicolumn{2}{c}{\textbf{23.8}}&\multicolumn{2}{|c|}{\textbf{41.3}}
\\
\hline
\end{tabular}
}
\caption{The performances of different scene completion methods on the NYU Kinect dataset. The best results are in bold.}
\vspace{-4mm}
\label{table:NYUK}
\end{center}
\end{table*}
\begin{table*}
\begin{center}
\doublerulesep=0.5pt
\resizebox{0.9\textwidth}{!}
{
\begin{tabular}{|c|ccccc|c|c|c|c|c|c|c|c|c|c|c|c|c|c|c|c|c|c|c|c|c|c|c|c|c|c|c|c|c|c|c|c|c|c|c|c|c|c|c|c|c|c|c|c|c|c|c|c|c|c|c|c|c|c|c|c|c|c|c|}
\hline
\multicolumn{8}{|c|}{}&\multicolumn{6}{|c|}{scene completion}&\multicolumn{24}{|c|}{semantic scene completion}
\\
\hline
\multicolumn{4}{|c|}{Methods}&\multicolumn{4}{|c|}{Trained on}
&\multicolumn{2}{|c}{prec.}&\multicolumn{2}{c}{recall}&\multicolumn{2}{c|}{IoU}
&\multicolumn{2}{|c}{ceil.}&\multicolumn{2}{c}{floor}&\multicolumn{2}{c}{wall}
&\multicolumn{2}{c}{win.}&\multicolumn{2}{c}{chair}&\multicolumn{2}{c}{bed}
&\multicolumn{2}{c}{sofa}&\multicolumn{2}{c}{table}&\multicolumn{2}{c}{tvs}
&\multicolumn{2}{c}{furn.}&\multicolumn{2}{c}{objs.}&\multicolumn{2}{|c|}{avg.}
\\
\hline
\multicolumn{4}{|c|}{Zheng et al.~\cite{zheng2013beyond}}&\multicolumn{4}{|c|}{NYU}
&\multicolumn{2}{|c}{60.1}&\multicolumn{2}{c}{46.7}&\multicolumn{2}{c|}{34.6}
&\multicolumn{2}{|c}{--}&\multicolumn{2}{c}{--}&\multicolumn{2}{c}{--}
&\multicolumn{2}{c}{--}&\multicolumn{2}{c}{--}&\multicolumn{2}{c}{--}
&\multicolumn{2}{c}{--}&\multicolumn{2}{c}{--}&\multicolumn{2}{c}{--}
&\multicolumn{2}{c}{--}&\multicolumn{2}{c}{--}&\multicolumn{2}{|c|}{--}
\\
\multicolumn{4}{|c|}{Firman et al.~\cite{firman2016structured}}&\multicolumn{4}{|c|}{NYU}
&\multicolumn{2}{|c}{66.5}&\multicolumn{2}{c}{69.7}&\multicolumn{2}{c|}{50.8}
&\multicolumn{2}{|c}{--}&\multicolumn{2}{c}{--}&\multicolumn{2}{c}{--}
&\multicolumn{2}{c}{--}&\multicolumn{2}{c}{--}&\multicolumn{2}{c}{--}
&\multicolumn{2}{c}{--}&\multicolumn{2}{c}{--}&\multicolumn{2}{c}{--}
&\multicolumn{2}{c}{--}&\multicolumn{2}{c}{--}&\multicolumn{2}{|c|}{--}
\\
\hline
\multicolumn{4}{|c|}{SSCNet~\cite{song2017semantic}}&\multicolumn{4}{|c|}{NYU}
&\multicolumn{2}{|c}{75.0}&\multicolumn{2}{c}{92.3}&\multicolumn{2}{c|}{70.3}
&\multicolumn{2}{|c}{--}&\multicolumn{2}{c}{--}&\multicolumn{2}{c}{--}
&\multicolumn{2}{c}{--}&\multicolumn{2}{c}{--}&\multicolumn{2}{c}{--}
&\multicolumn{2}{c}{--}&\multicolumn{2}{c}{--}&\multicolumn{2}{c}{--}
&\multicolumn{2}{c}{--}&\multicolumn{2}{c}{--}&\multicolumn{2}{|c|}{--}
\\
\multicolumn{4}{|c|}{SSCNet~\cite{song2017semantic}}&\multicolumn{4}{|c|}{NYU+SUNCG}
&\multicolumn{2}{|c}{75.4}&\multicolumn{2}{c}{\textbf{96.3}}&\multicolumn{2}{c|}{73.2}
&\multicolumn{2}{|c}{32.5}&\multicolumn{2}{c}{92.6}&\multicolumn{2}{c}{40.2}
&\multicolumn{2}{c}{8.9}&\multicolumn{2}{c}{33.9}&\multicolumn{2}{c}{57.0}
&\multicolumn{2}{c}{59.5}&\multicolumn{2}{c}{28.3}&\multicolumn{2}{c}{8.1}
&\multicolumn{2}{c}{44.8}&\multicolumn{2}{c}{25.1}&\multicolumn{2}{|c|}{40.0}
\\
\multicolumn{4}{|c|}{VVNet~\cite{guo2018view}}&\multicolumn{4}{|c|}{NYU+SUNCG}
&\multicolumn{2}{|c}{86.4}&\multicolumn{2}{c}{92.0}&\multicolumn{2}{c|}{80.3}
&\multicolumn{2}{|c}{--}&\multicolumn{2}{c}{--}&\multicolumn{2}{c}{--}
&\multicolumn{2}{c}{--}&\multicolumn{2}{c}{--}&\multicolumn{2}{c}{--}
&\multicolumn{2}{c}{--}&\multicolumn{2}{c}{--}&\multicolumn{2}{c}{--}
&\multicolumn{2}{c}{--}&\multicolumn{2}{c}{--}&\multicolumn{2}{|c|}{--}
\\
\multicolumn{4}{|c|}{DCRF~\cite{zhang2018semantic}}&\multicolumn{4}{|c|}{NYU}
&\multicolumn{2}{|c}{--}&\multicolumn{2}{c}{--}&\multicolumn{2}{c|}{--}
&\multicolumn{2}{|c}{35.5}&\multicolumn{2}{c}{92.6}&\multicolumn{2}{c}{52.4}
&\multicolumn{2}{c}{10.7}&\multicolumn{2}{c}{40.0}&\multicolumn{2}{c}{60.0}
&\multicolumn{2}{c}{62.5}&\multicolumn{2}{c}{34.0}&\multicolumn{2}{c}{9.4}
&\multicolumn{2}{c}{49.2}&\multicolumn{2}{c}{26.5}&\multicolumn{2}{|c|}{43.0}
\\
\multicolumn{4}{|c|}{TS3D,V2~\cite{garbade2018two}}&\multicolumn{4}{|c|}{NYU}
&\multicolumn{2}{|c}{81.2}&\multicolumn{2}{c}{93.6}&\multicolumn{2}{c|}{76.9}
&\multicolumn{2}{|c}{33.9}&\multicolumn{2}{c}{93.4}&\multicolumn{2}{c}{47.0}
&\multicolumn{2}{c}{26.4}&\multicolumn{2}{c}{27.9}&\multicolumn{2}{c}{61.7}
&\multicolumn{2}{c}{51.7}&\multicolumn{2}{c}{27.6}&\multicolumn{2}{c}{27.3}
&\multicolumn{2}{c}{44.4}&\multicolumn{2}{c}{21.8}&\multicolumn{2}{|c|}{42.1}
\\
\multicolumn{4}{|c|}{TS3D,V3+~\cite{garbade2018two}}&\multicolumn{4}{|c|}{NYU}
&\multicolumn{2}{|c}{80.2}&\multicolumn{2}{c}{94.4}&\multicolumn{2}{c|}{76.5}
&\multicolumn{2}{|c}{34.4}&\multicolumn{2}{c}{93.6}&\multicolumn{2}{c}{47.7}
&\multicolumn{2}{c}{31.8}&\multicolumn{2}{c}{32.2}&\multicolumn{2}{c}{65.2}
&\multicolumn{2}{c}{54.2}&\multicolumn{2}{c}{30.7}&\multicolumn{2}{c}{32.5}
&\multicolumn{2}{c}{50.1}&\multicolumn{2}{c}{30.7}&\multicolumn{2}{|c|}{45.7}
\\
\multicolumn{4}{|c|}{DDRNet~\cite{li2019rgbd}}&\multicolumn{4}{|c|}{NYU}
&\multicolumn{2}{|c}{88.7}&\multicolumn{2}{c}{88.5}&\multicolumn{2}{c|}{79.4}
&\multicolumn{2}{|c}{54.1}&\multicolumn{2}{c}{91.5}&\multicolumn{2}{c}{56.4}
&\multicolumn{2}{c}{14.9}&\multicolumn{2}{c}{37.0}&\multicolumn{2}{c}{55.7}
&\multicolumn{2}{c}{51.0}&\multicolumn{2}{c}{28.8}&\multicolumn{2}{c}{9.2}
&\multicolumn{2}{c}{44.1}&\multicolumn{2}{c}{27.8}&\multicolumn{2}{|c|}{42.8}
\\
\multicolumn{4}{|c|}{Ours}&\multicolumn{4}{|c|}{NYU}
&\multicolumn{2}{|c}{91.3}&\multicolumn{2}{c}{92.6}&\multicolumn{2}{c|}{82.4}
&\multicolumn{2}{|c}{56.2}&\multicolumn{2}{c}{94.6}&\multicolumn{2}{c}{58.7}
&\multicolumn{2}{c}{35.1}&\multicolumn{2}{c}{44.8}&\multicolumn{2}{c}{68.6}
&\multicolumn{2}{c}{65.3}&\multicolumn{2}{c}{37.6}&\multicolumn{2}{c}{35.5}
&\multicolumn{2}{c}{53.1}&\multicolumn{2}{c}{35.2}&\multicolumn{2}{|c|}{53.2}
\\
\multicolumn{4}{|c|}{Ours}&\multicolumn{4}{|c|}{NYU+SUNCG}
&\multicolumn{2}{|c}{\textbf{93.4}}&\multicolumn{2}{c}{91.2}&\multicolumn{2}{c|}{\textbf{85.1}}
&\multicolumn{2}{|c}{\textbf{58.1}}&\multicolumn{2}{c}{\textbf{95.1}}&\multicolumn{2}{c}{\textbf{60.5}}
&\multicolumn{2}{c}{\textbf{36.8}}&\multicolumn{2}{c}{\textbf{47.2}}&\multicolumn{2}{c}{\textbf{69.3}}
&\multicolumn{2}{c}{\textbf{67.7}}&\multicolumn{2}{c}{\textbf{39.8}}&\multicolumn{2}{c}{\textbf{37.6}}
&\multicolumn{2}{c}{\textbf{55.4}}&\multicolumn{2}{c}{\textbf{37.6}}&\multicolumn{2}{|c|}{\textbf{55.0}}
\\
\hline
\end{tabular}
}
\caption{The performances of different scene completion methods on the NYU CAD dataset. The best results are in bold.}
\label{table:NYUCAD}
\vspace{-6mm}
\end{center}
\end{table*}
\subsubsection{Comparison on the SUNCG dataset.}
For the SUNCG dataset, we compare our proposed CCPNet with SSCNet~\cite{song2017semantic}, VVNet~\cite{guo2018view}, DCRF~\cite{zhang2018semantic}, ESSCNet~\cite{zhang2018efficient} and SATNet~\cite{liu2018see} for both SC and SSC tasks.
As shown in Tab.~\ref{table:SUNCG}, our approach achieves the best performance in both SC and SSC tasks.
Compared to the SSCNet, the overall IoUs of our CCPNet significantly increase about 18\% and 28\% for SC and SSC tasks, respectively.
In spite of taking a single depth map, our approach gets higher IoUs than the RGB-D based SATNet (Ours 91.4\% vs. SATNet 78.5\%).
Our approach also perform better than the previous best ESSCNet with a considerable margin.
Tab.~\ref{table:SUNCG} also lists the IoU for each object category.
Our approach also achieves the highest IoUs in each category.
Thus, the quantitative results demonstrate that our approach is superior in 3D SSC.
Fig.~\ref{fig:SUNCG} illustrates the qualitative results on the SUNCG dataset.
Although previous methods works well for many scenes, they usually fail in the objects which have complex structures and confusing semantics (the first and second rows).
In contrast, our method leverages the low-level features and multi-scale contexts to overcome these difficulties.
\begin{figure*}
\centering
\resizebox{1\textwidth}{!}
{
\begin{tabular}{@{}c@{}c@{}c@{}c@{}c@{}c@{}c@{}c@{}c@{}c@{}c@{}c@{}c@{}c@{}c@{}c@{}c@{}c@{}c}
(a)\hspace{2.2cm}(b)\hspace{2cm}(c)\hspace{2cm}(d)\hspace{2cm}(e)\hspace{2cm}(f)\hspace{2.2cm}(g)\\
\includegraphics[width=1\linewidth,height=4cm]{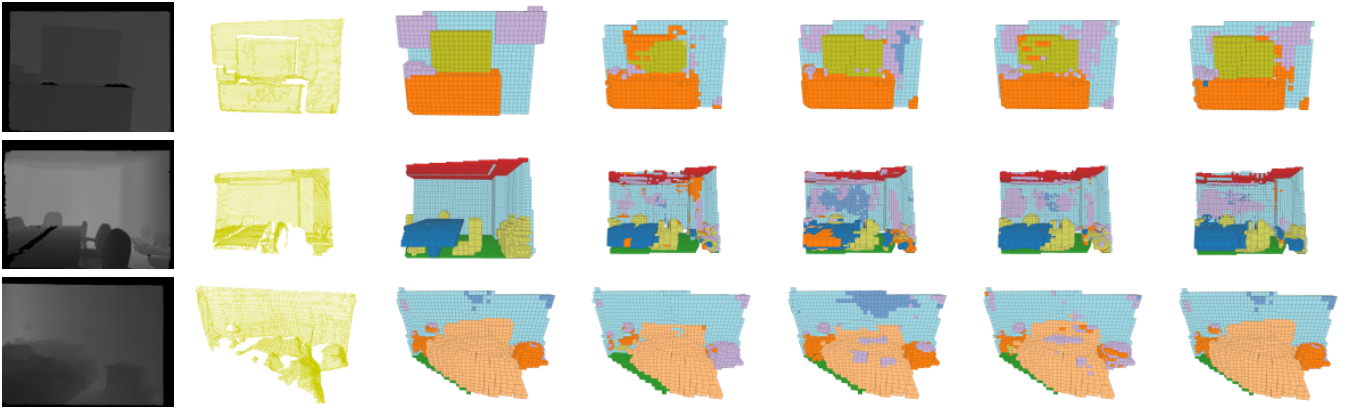} \\
\includegraphics[width=0.8\linewidth,height=0.5cm]{legend.pdf} \\
\end{tabular}
}
\caption{Completion results with different methods on the NYU dataset.
From the left to right: (a) Input Depth; (b) fTSDF Surface; (c) Ground Truth; (d) SSCNet~\cite{song2017semantic}; (e) DDRNet~\cite{li2019rgbd}; (f) VVNet~\cite{guo2018view}; (g) Ours. The figure is best viewed in color with 200\% zooming-in.}
\vspace{-4mm}
\label{fig:NYU}
\end{figure*}
\vspace{-2mm}
\subsubsection{Comparison on the NYU dataset.}
For the NYU dataset, we compare our CCPNet with other outstanding methods.
Tab.~\ref{table:NYUK} and Tab.~\ref{table:NYUCAD} illustrate the performances on the NYU Kinect and NYU CAD datasets, respectively.
From the results, we can see that our CCPNet also achieves the best performance.
For the SC task, it outperforms the SSCNet (8.4\% on NYU Kinect and 12.1\% on NYU CAD) when only the NYU dataset is used as the training data.
Meanwhile, even the SSCNet uses the additional SUNCG training dataset, our CCPNet still achieves a substantial improvement (7\% on NYU Kinect and 9.2\% on NYU CAD).
We observe that the SSCNet achieves a rather high recall but a low precision for the SC task.
Our model pre-trained with the SUNCG dataset achieves better performances, and outperforms previous best methods, \emph{i.e.}, VVNet and SATNet, with a large margin.

For the SSC task, our approach achieves 41.3\% on NYU Kinect and 55.0\% on NYU CAD, and outperforms the SSCNet~\cite{song2017semantic} by 10.8\% and 15\%, respectively.
With the same training data, our approach constantly performs better than existing best methods with a considerable margin.
Tab.~\ref{table:NYUK} and Tab.~\ref{table:NYUCAD} also include the results of each category.
In general, our method tends to predict more occluded voxels than previous methods, such as window, chair and furniture. %
Fig.~\ref{fig:NYU} shows the qualitative results in which cluttered scene completions can be observed.
Our method performs substantially better than other approaches.
\subsubsection{Efficiency Analysis}
Current methods usually depend on expensive 3D CNNs and feature concatenations, while our CCPNet utilizes a light-weight 3D dilated encoder and a self-cascaded pyramid.
Thus, it significantly reduces memory requirement and computational cost for inference.
Tab.~\ref{table:computation} lists the parameters and computations of different methods.
Our CCPNet achieves much better accuracy, and significantly reduces the model parameters, and speeds up for inference.
\subsection{Ablation Studies}
To verify the effect of our proposed modules, we also perform ablation experiments on
the SUNCG dataset.
\begin{table}
\begin{center}
\doublerulesep=0.5pt
\resizebox{0.48\textwidth}{!}
{
\begin{tabular}{|c|c|c|c|c|c|c}
\hline
Methods &Params/k&FLOPs/G&Speed/ms&SC-IoU&SSC-IoU\\
\hline
\hline
SSCNet~\cite{song2017semantic}&930&163.8&578& 55.1& 24.7 \\
VVNet~\cite{guo2018view}&685&119.2&74& 61.1& 32.9\\
ESSCNet~\cite{zhang2018efficient}&160&22.0&121& 56.2&26.7\\
SATNet~\cite{liu2018see}&1200&187.5&1300& 60.6& 34.4\\
DDRNet~\cite{li2019rgbd}&195&27.2&658& 61.0& 30.4\\
Ours&\textbf{89}&\textbf{11.8}&\textbf{57}& \textbf{67.1}& \textbf{41.3}\\
\hline
\end{tabular}
}
\caption{Comparison of efficiency with different methods.}
\label{table:computation}
\vspace{-4mm}
\end{center}
\end{table}

\textbf{Separated Convolution Kernels.}
Based on the SSCNet~\cite{song2017semantic}, we replace the 3D dilated convolutions of SSCNet with our proposed separated kernels.
For simplification, we set the number of subvolumes to 4.
Tab.~\ref{table:sk} shows the quantitative performances.
For SC and SSC tasks, our method has fewer parameters and computations, while provides 3.3\% and 6.1\% IoU improvements compared to the SSCNet.
\begin{table}
\begin{center}
\doublerulesep=0.5pt
\resizebox{0.48\textwidth}{!}
{
\begin{tabular}{|c|c|c|c|c|cc}
\hline
Methods &SC-IoU&SSC-IoU&Params/k&FLOPs/G\\
\hline
\hline
SSCNet~\cite{song2017semantic}&73.5& 46.4& 930& 163.8\\
SSCNet~\cite{song2017semantic}+SK&\textbf{76.8}&\textbf{52.5}&\textbf{532}&\textbf{100.3}\\
\hline
\end{tabular}
}
\caption{Quantitative results on separated convolution kernels.}
\label{table:sk}
\vspace{-6mm}
\end{center}
\end{table}

\textbf{Cascaded Context Pyramid.}
To verify the effect of our CCP, we replace the CCP with the outstanding PPM~\cite{zhao2017pyramid} and ASPP~\cite{chen2018deeplab} modules, and keep other modules unchanged.
The first three rows of Tab.~\ref{table:abs} show the quantitative results.
With the PPM and ASPP, the IoUs of the CCPNet decrease 4.1\% and 2.3\% for the SC task, respectively.
For the SSC task, it has a similar trend, which proves that our CCP is more effective.
Note that the PPM and ASPP need more memories and parameters for the context aggregation.

\textbf{Guided Residual Refinement.}
%
To evaluate the effect of our GRR, we compare the performances with different refinements.
As shown in the 4-th row of Tab.~\ref{table:abs}, with the BRBs, the CCPNet shows worse results, decreasing 8.1\% and 8.4\% for SC and SSC, respectively.
However, when introducing the guidance (the 5-th row), the model shows significant improvements for both SC and SSC tasks.
Only with the feature amplification (the 6-th row), we observe a considerable improvement compared to the BRBs.
A possible reason is that it is not enough for the detail recovery when only amplifying on the 3D context information. However, with the whole GRB, our approach shows best results.
\begin{table}
\begin{center}
\doublerulesep=0.5pt
\resizebox{0.48\textwidth}{!}
{
\begin{tabular}{|c|c|c|c|c|c|c|}
\hline
Methods &SC-IoU&SSC-IoU&Params/k&FLOPs/G\\
\hline
\hline
CCPNet&\textbf{91.4} & \textbf{74.2}&89 & 11.8 \\
CCPNet (CCP$\to$ PPM)&87.3&71.6&120&87.2\\
CCPNet (CCP$\to$ ASPP)&89.1&72.3&145&140.2\\
\hline
CCPNet (GRB$\to$BRB)&83.3&65.8&89 & 9.2\\
CCPNet (GRB w/o Ampli)&88.7& 73.6&89 &11.5 \\
CCPNet (GRB w/o Guidance)&84.3&67.4&89 &11.2 \\
\hline
CCPNet-Quarter &86.5&69.1&\textbf{76}& \textbf{6.5}\\
CCPNet-Half&88.4&73.1&81& 10.4\\
\hline
\end{tabular}
}
\caption{Ablation results of components on the SUNCG dataset.}
\label{table:abs}
\vspace{-6mm}
\end{center}
\end{table}

\vspace{-4mm}
\textbf{Full-Resolution Prediction.}
To evaluate the benefits of full-resolutions, we also re-implement our approach with the quarter and half resolution.
To achieve this goal, we remove the corresponding layers after the deconvolution operations in Fig.~\ref{fig:framework}.
The last two rows of Tab.~\ref{table:abs} show the performances.
From the results, we can see that the low-resolution-based model shows worse performances.
The main reason is that it cannot preserve the geometric details.
However, our model still performs better than most state-of-the-art methods.
This further demonstrates the effectiveness of our proposed modules.
With full-resolution outputs, our model can fully exploit the geometric details, improving the IoUs by 4.9\% and 5.1\% respectively.
\section{Conclusion}
In this work, we propose a novel deep learning framework, named CCPNet, for full-resolution 3D SSC.
The CCPNet is a self-cascaded pyramid structure to successively aggregate multi-scale 3D contexts and local geometry details.
Extensive experiments on both synthetic and real benchmarks demonstrate that our CCPNet significantly improves the semantic completion accuracy, reduces the computational cost, and offers high-quality completion results with full-resolution.
In the future work, we will explore color information for semantic and boundary enhancement.

\textbf{Acknowledgements.}
This work is partly supported by the National Natural Science Foundation of China (No. 61725202, 61751212 and 61829102), the Key Research and Development Program of Sichuan Province (No. 2019YFG0409), and the Fundamental Research Funds for the Central Universities (No. DUT19GJ201).

{\small
\bibliographystyle{ieee}
\bibliography{egbib}
}

\end{document}